\documentclass{article}

\usepackage[utf8]{inputenc} 
\usepackage[T1]{fontenc}   
\usepackage{times}          

\usepackage{amsmath}
\usepackage{amsfonts}
\usepackage{nicefrac}    

\usepackage[pdftex]{graphicx}
\graphicspath{{./pdf/}}
\usepackage{adjustbox}
\usepackage{placeins}

\usepackage{booktabs}   
\usepackage{multirow}
\usepackage{tabularx}

\usepackage{pdflscape}

\usepackage{enumitem}
\usepackage{listings}
\usepackage{microtype}    

\usepackage[dvipsnames]{xcolor}

\usepackage[numbers]{natbib}
\usepackage{doi}
\usepackage{xurl}        
\usepackage{url}
\usepackage{hyperref}     

\usepackage{csagh}       
\makeatletter
\def\enditemize{%
  \endlist
  \global\@itemdepth\z@ 
}
\def\endenumerate{%
  \endlist
  \global\@enumdepth\z@ 
}
\makeatother

\usepackage{lipsum}        

\graphicspath{{./pdf/}}

\begin{document}
\begin{opening}

\title{Bielik 11B v3: Multilingual Large Language Model for European Languages}

\author[SpeakLeash, Azurro, krzysztof.ociepa@bielik.ai]{Krzysztof Ociepa}
\author[SpeakLeash, ACK Cyfronet AGH, lukasz.flis@cyfronet.pl]{\L{}ukasz Flis}
\author[SpeakLeash, remigiusz.kinas@bielik.ai]{Remigiusz Kinas}
\author[SpeakLeash, Jagiellonian University, Enelpol, krzysztof.pawel.wrobel@uj.edu.pl]{Krzysztof Wr\'obel}
\author[SpeakLeash, ACK Cyfronet AGH, adrian.gwozdziej@cyfronet.pl]{Adrian Gwo\'zdziej}

\received{TODO}
\revised{TODO}
\accepted{TODO}

\begin{abstract}
We present Bielik 11B v3, a state-of-the-art language model highly optimized for the Polish language, while also maintaining strong capabilities in other European languages. This model extends the Mistral 7B v0.2 architecture, scaled to 11B parameters via depth up-scaling. Its development involved a comprehensive four-stage training pipeline: continuous pre-training, supervised fine-tuning (SFT), Direct Preference Optimization (DPO), and reinforcement learning.

Comprehensive evaluations demonstrate that Bielik 11B v3 achieves exceptional performance. It significantly surpasses other specialized Polish language models and outperforms many larger models (with 2-6 times more parameters) on a wide range of tasks, from basic linguistic understanding to complex reasoning.

The model's parameter efficiency, combined with extensive quantization options, allows for effective deployment across diverse hardware configurations. Bielik 11B v3 not only advances AI capabilities for the Polish language but also establishes a new benchmark for developing resource-efficient, high-performance models for less-represented languages.
\end{abstract}



\end{opening}

\section*{Note} 
This is a preliminary version; results and exposition may change.

\section{Introduction}

The development of large language models for underrepresented languages has gained significant momentum, with several initiatives targeting European languages. Our work builds upon our previous Bielik 11B v2 model \cite{ociepa2025bielik11bv2technical} and complements our smaller Bielik v3 models \cite{ociepa2025bielikv3smalltechnical}. This effort aligns with broader European initiatives, including EuroLLM \cite{MARTINS202553}, which targets multilingual capabilities across EU languages, Apertus \cite{swissai2025apertus}, focused on democratizing open LLMs for global language environments, and Polish-specific projects such as the PLLuM family \cite{kocon2025pllumfamilypolishlarge}.

\section{Model and Tokenizer}
This section provides an overview of the model architecture and tokenizer setup, explaining the main design principles and implementation configurations.

\subsection{Model Architecture}

\begin{table}[h]
\centering
\begin{tabular}{ll}
\toprule
Parameter & Value \\
\midrule
Layers & 50 \\
Model Dimension & 4096 \\
Attention Heads & 32 \\
Key/Value Heads & 8 \\
Head Size & 128 \\
Intermediate Size & 14336 \\
Activation Function & SwiGLU \\
Vocabulary Size & 32128 \\
Positional Embeddings & RoPE ($\theta = 1000000$) \\
Native Context Length & 32768 \\
Extended Context Length (YaRN) & 131072 \\
\bottomrule
\end{tabular}
\caption{Model architecture.}
\label{tab:model-architecture}
\end{table}

\noindent The Bielik 11B v3 architecture is founded on the Transformer model \cite{Vaswani2017AttentionIA}, with its principal configuration parameters summarized in Table \ref{tab:model-architecture}. The design integrates a suite of contemporary architectural and algorithmic improvements intended to enhance computational efficiency, stability, and overall model performance.

\noindent \textbf{Self-attention with causal masking} \cite{Vaswani2017AttentionIA} enables the network to dynamically allocate attention across different elements of the input sequence. The application of a causal mask constrains attention to preceding tokens, thereby preserving the autoregressive property critical for effective language modeling.

\noindent \textbf{Grouped-query attention (GQA)} \cite{ainslie-etal-2023-gqa} provides a more resource-efficient variant of multi-head attention by employing fewer key–value heads than query heads. This configuration substantially reduces both computational and memory overheads while maintaining comparable representational fidelity, thus facilitating more efficient processing of extended input sequences.

\noindent \textbf{SwiGLU activation} \cite{Dauphin2016LanguageMW,shazeer2020gluvariantsimprovetransformer} combines the Swish nonlinearity with Gated Linear Units (GLU), offering improved gradient flow and optimization stability. Empirical evidence indicates that SwiGLU yields superior performance and trainability relative to conventional activation functions such as ReLU.

\noindent \textbf{Rotary Positional Embeddings (RoPE)} \cite{SU2024127063} are employed to encode relative positional information within token representations. In contrast to absolute positional embeddings, RoPE provides enhanced generalization to longer sequences and improved modeling of positional dependencies, particularly in tasks sensitive to word order and spatial context.

\noindent \textbf{Root Mean Square Layer Normalization (RMSNorm)} \cite{10.5555/3666122.3668105} normalizes activations based on their root mean square, rather than their mean and variance as in standard Layer Normalization. This approach yields marginally faster computation and greater numerical stability during training, contributing to overall model robustness.

\noindent \textbf{Pre-normalization} applies layer normalization prior to the self-attention and feed-forward sublayers. This architectural choice enhances gradient flow throughout the network, resulting in improved convergence behavior and more consistent model performance.

Building upon the substantial success of the previous Bielik 11B v2 \cite{ociepa2025bielik11bv2technical} iteration, we adopted an identical, validated architectural approach. This strategic decision to leverage an existing, high-performing model structure, rather than commencing development from scratch, was fundamentally motivated by the imperative for efficient resource allocation. By concentrating our efforts on the linguistic and structural refinement of a proven base model, we achieved significant optimization in both time and computational resources.

\begin{figure}[ht]
\centering
\includegraphics[width=\columnwidth]{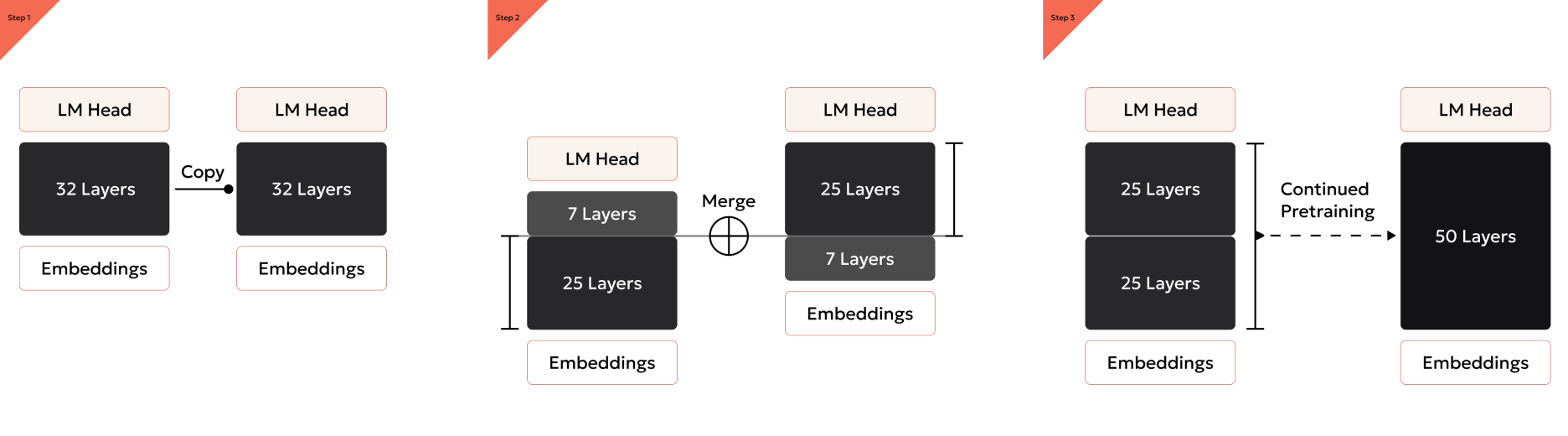}
\caption{Depth up-scaling with $n = 32$, $m = 7$, and $s = 50$.}
\label{fig:model-upscaling}
\end{figure}

The resulting Bielik 11B v3 model is an adaptation of the Mistral 7B v0.2 model \cite{jiang2023mistral7b}, meticulously scaled via the Depth Up-Scaling methodology \cite{kim2024solar107bscalinglarge}, followed by a subsequent phase of continued pretraining. Specifically, we initiated the process with the original $n = 32$-layer architecture. The up-scaling procedure involved duplicating the model's existing layers, followed by the strategic excision of the final eight and initial eight layers at the central junction ($m=7$), culminating in a $s = 50$-layer model, as schematically detailed in Figure \ref{fig:model-upscaling}. This target of 50 layers was chosen to achieve a final model size of approximately 11 billion parameters, which retains the crucial operational characteristic of being runnable on mainstream, consumer-grade GPUs equipped with up to 24 GB of VRAM.

\subsection{Tokenizer}
\begin{table*}[h]
\centering
\begin{tabular}{lll|lll|lll}
\toprule
                          &  Vocab               &  Avg              & \multicolumn{3}{c|}{Polish} & \multicolumn{3}{c}{English} \\
Tokenizer                 &  size &  tokens & Tokens   & CpT    & TpW    & Tokens    & CpT    & TpW    \\
EuroLLM 9B                & 128000          & 420            & 437      & 4.11   & 1.88   & 404       & 4.79   & 1.27   \\
Mistral Small 3.2 24B     & 131072          & 462            & 547      & 3.28   & 2.36   & 377       & 5.14   & 1.19   \\
APT3                      & 31980           & 480            & 344      & 5.22   & 1.48   & 615       & 3.15   & 1.93   \\
Qwen3                     & 151669          & 499            & 625      & 2.87   & 2.69   & 373       & 5.19   & 1.17   \\
APT4                      & 32000           & 503            & 375      & 4.78   & 1.62   & 631       & 3.07   & 1.98   \\
Llama2                    & 32000           & 554            & 681      & 2.63   & 2.94   & 427       & 4.53   & 1.34   \\
Mistral 7B v0.2           & 32000           & 578            & 747      & 2.40   & 3.22   & 408       & 4.75   & 1.28   \\
\bottomrule
\end{tabular}
\caption{Comparison of token count, characters per token (CpT), and tokens per word (TpW) for the preamble of the Constitution of the Republic of Poland in Polish and English versions, processed by various tokenizers.}
\label{tab:tokenizers-comparison}
\end{table*}

The tokenization efficiency is primarily assessed by the resulting token count for a given input sequence, where a reduction in this number typically correlates with faster and more resource-efficient text generation by the language model. The original tokenizer from the Mistral 7B v0.2 model, however, was not specifically trained on or optimized for the Polish language.

To establish a relevant baseline for comparative analysis, we selected the Preamble of the Constitution of the Republic of Poland as our benchmark text. This text was chosen for its comprehensive representation of formal Polish writing, and critically, because it possesses an official English translation, allowing for direct comparison. Table \ref{tab:tokenizers-comparison} provides a detailed comparison of key metrics, including token count, characters per token (CpT), and tokens per word (TpW), to illustrate the performance of different tokenizers when applied to both the Polish and English versions of the preamble.

Ultimately, we elected to retain the original tokenizer derived from the Mistral 7B v0.2 model, which encompasses a base vocabulary of 32,000 tokens. The sole modification introduced was the necessary addition of special tokens required for the standardized chat template, increasing the total vocabulary size to 32,128 tokens, which aligns precisely with the tokenizer used in the preceding Bielik 11B v2.6 version \cite{Bielik11Bv26i}. The rationale for maintaining this specific tokenizer was twofold: first, the Bielik 11B v3 model is designed to support over 30 languages; and second, the current tokenizer provides a satisfactory trade-off between tokenization efficiency for Polish and its performance across other supported languages, including English, while maintaining a relatively small vocabulary size.

\section{Pre-training}

The central objective of the pre-training phase was to substantially enhance the model's linguistic proficiency, specifically targeting high levels of accuracy and fluency across Polish and other major European languages. To realize this goal, a heterogeneous corpus comprising high-quality textual data was employed. These materials underwent a rigorous preprocessing pipeline and comprehensive quality evaluation to ensure optimal standards for the training data utilized in the subsequent fine-tuning processes.

\subsection{Pre-training Data} \label{Pre-training-Data}

\begin{table*}[h]
\centering
\begin{tabular}{|l|r|c|}
\hline
\textbf{Language} & \textbf{Number of documents} & \textbf{Share of dataset} \\ \hline
Polish & 428.6M & 54.25\% \\ \hline
English & 162.0M & 20.50\% \\ \hline
Dutch & 12.8M & 1.62\% \\ \hline
Portuguese & 12.2M & 1.55\% \\ \hline
Swedish & 11.9M & 1.50\% \\ \hline
German & 11.5M & 1.45\% \\ \hline
Spanish & 10.7M & 1.36\% \\ \hline
Italian & 10.3M & 1.31\% \\ \hline
Turkish & 9.7M & 1.22\% \\ \hline
French & 9.7M & 1.22\% \\ \hline
Czech & 8.6M & 1.09\% \\ \hline
Romanian & 8.6M & 1.09\% \\ \hline
Bosnian & 8.4M & 1.06\% \\ \hline
Finnish & 8.1M & 1.02\% \\ \hline
Hungarian & 7.5M & 0.95\% \\ \hline
Greek & 6.5M & 0.82\% \\ \hline
Ukrainian & 6.2M & 0.79\% \\ \hline
Slovenian & 6.2M & 0.78\% \\ \hline
Danish & 6.0M & 0.76\% \\ \hline
Russian & 5.9M & 0.74\% \\ \hline
Croatian & 5.9M & 0.74\% \\ \hline
Bulgarian & 5.6M & 0.71\% \\ \hline
Slovak & 5.6M & 0.71\% \\ \hline
Lithuanian & 5.2M & 0.66\% \\ \hline
Norwegian & 5.2M & 0.65\% \\ \hline
Estonian & 3.9M & 0.49\% \\ \hline
Latvian & 3.6M & 0.46\% \\ \hline
Albanian & 1.1M & 0.14\% \\ \hline
Serbian & 1.1M & 0.14\% \\ \hline
Belarusian & 920.1K & 0.12\% \\ \hline
Icelandic & 759.4K & 0.10\% \\ \hline
Serbo-Croatian & 84.0K & 0.01\% \\ \hline
\end{tabular}
\caption{Language distribution of documents used in pretraining the Bielik 11B v3 model}
\label{tab:b3_pretraining_data_langs}
\end{table*}

The pre-training phase for Bielik v3 involved a substantial expansion in both the volume and variety of training corpora compared to its predecessor, Bielik 11B v2 \cite{ociepa2025bielik11bv2technical}. Specifically, the total token count was increased fivefold, encompassing a multilingual corpus across 32 languages (detailed in Table \ref{tab:b3_pretraining_data_langs}). This newly curated dataset integrates a broad spectrum of high-quality sources, including:
\begin{itemize}
    \item \textbf{Legal and Official Records:} Court rulings, legislative acts, regulations, and official resolutions.
    \item \textbf{Academic and Press:} Scientific literature from the Science Library and commercial press publications.
    \item \textbf{Public Discourse:} Parliamentary proceedings and thematic web forums.
    \item \textbf{Cultural and Digital Repositories:} Content from Wikipedia, Europeana, and public web domains.
    \item \textbf{Synthetic Data:} Purpose-built datasets generated to augment training.
\end{itemize}

A primary emphasis was placed on Polish and European contexts. This includes specialized efforts to capture regional nuances within Poland, such as the integration of Silesian and Kashubian Wikipedia editions. The collection also draws from established repositories like the Parliamentary Discourse Corpus (\textit{Korpus Dyskursu Parlamentarnego}), the Science Library (\textit{Biblioteka Nauki}), and various anonymized journals of laws and court gazettes.

\subsubsection{Data Processing and Quality Control}

To ensure the integrity of the model, we implemented a rigorous preprocessing pipeline. While the corpus includes copyright-protected materials and documents that may contain personal data, a robust anonymization mechanism is standard for all pipelines. This system identifies and removes sensitive identifiers, including PESEL numbers, phone numbers, email addresses, and URLs. 

Furthermore, we applied deduplication and quality-filtering heuristics to minimize redundancy. These measures are applied proportionally, adjusting for the specific licensing terms and the nature of the data source (e.g., API-based access versus public dumps). Detailed methodologies for these filtering processes are documented in the technical reports for our previous Bielik 11B v2 model \cite{ociepa2025bielik11bv2technical} and our smaller Bielik v3 models \cite{ociepa2025bielikv3smalltechnical}.

\subsubsection{Ethical Compliance and Data Rights}

Our curation process adheres to strict ethical and legal standards. We systematically monitor \texttt{robots.txt} files, meta-tag-based Text and Data Mining (TDM) reservations, and Terms of Service (ToS) declarations. Any source expressing a reservation of rights is excluded from the corpus. 

This exclusion policy is applied retroactively: should a data provider declare a reservation after initial collection, the relevant data is purged from all future iterations of the model. This protocol extends to our own publicly released datasets; if a manifest explicitly prohibits AI training, those datasets are excluded and the restriction is publicly recorded.

\subsubsection{Linguistic Distribution and Sources}

The dataset is predominantly Polish and English, supplemented by other European languages. The collection effort, which commenced in November 2022, remains an ongoing initiative. The final training mix involves a sophisticated blending of various web domains, further refined through deduplication. Key external resources utilized include:
\begin{itemize}
    \item \textbf{CulturaX} \cite{nguyen-etal-2024-culturax} and \textbf{HPLT v2.0} \cite{burchell2025expandedmassivemultilingualdataset} for multilingual depth.
    \item \textbf{FineWeb} \cite{penedo2025fineweb2pipelinescale} and \textbf{FineWeb-Edu} \cite{lozhkov2024fineweb-edu} for high-quality web content.
    \item \textbf{SlimPajama-627B} \cite{cerebras2023slimpajama} and \textbf{Common Crawl}\footnote{\href{https://commoncrawl.org/}{https://commoncrawl.org/}} for large-scale general knowledge.
\end{itemize}

\subsection{Pre-training Stages}
The Bielik 11B v3 model was pre-trained through a three-stage process:

\begin{enumerate}
    \item \textbf{General Stage:} In the first pre-training stage, the model was trained on over $1$ trillion tokens using a context length of $8,192$ tokens. During this stage, the models achieved full pre-training on language proficiency and general world knowledge, with training data covering multiple languages.
    
    \item \textbf{Full Context Stage:} To increase the context length to the full size of $32,768$ tokens, we optimized the pre-training corpus for this stage by increasing the proportion of long documents. Specifically, we selected only documents with a length exceeding $7,168$ tokens. The models were further pre-trained with approximately $50$ billion tokens at a context length of $32,768$ tokens. We also increased the learning rate at this stage back to the initial value of $2.5 \times 10^{-5}$.
    
    \item \textbf{Long Context Stage:} In the final pre-training stage, we introduced YARN \cite{yarn_ICLR2024_874a4d89} and collected a high-quality long context corpora to extend the model's context length. The model was trained on over $1$ billion tokens using a maximum context length of $65,536$ tokens.
\end{enumerate}

At the conclusion of each stage, we merged selected recent checkpoints, which helped stabilize the model's performance. The selection process involved choosing checkpoints that achieved the highest scores on evaluation benchmarks. We then created various combinations of these selected checkpoints for merging, subsequently evaluated the merged models using the same benchmarks, and finally chose the best-performing one for continuation of the training process.

\subsection{Pre-training Hyperparameters}

The optimization process for the model was driven by the AdamW optimizer \cite{Loshchilov2017DecoupledWD}. We configured the optimizer with decay coefficients $\beta_1 = 0.9$ and $\beta_2 = 0.95$, alongside a weight decay coefficient of $0.1$. To regulate the learning rate, we implemented a cosine decay schedule that peaked at $2.5 \times 10^{-5}$ and tapered to a minimum of $9 \times 10^{-6}$. This schedule was preceded by a linear warm-up phase spanning the initial $50$ iterations. The entire training run was executed over a total of $270,000$ iterations, processing an aggregate of $1.1$ trillion tokens.

The training was divided into distinct phases to manage computational efficiency and context scaling:
\begin{itemize}
    \item \textbf{Baseline Phase:} Initial training was performed with a sequence length of $8,192$ tokens and a global batch size of $256$. During this stage, tensor parallelism was not utilized.
    \item \textbf{Context Expansion Phase:} To extend the model's capabilities to $32,768$ and $65,536$ tokens, we enabled tensor parallelism and integrated gradient checkpointing to manage memory constraints.
\end{itemize}

Stability and efficiency were further ensured by setting a gradient clipping threshold of $1.0$ and utilizing mixed-precision training with the $\text{bfloat}16$ format.

\subsection{Pre-training Process Monitoring}

\begin{figure}[ht]
\centering
\includegraphics[width=\columnwidth]{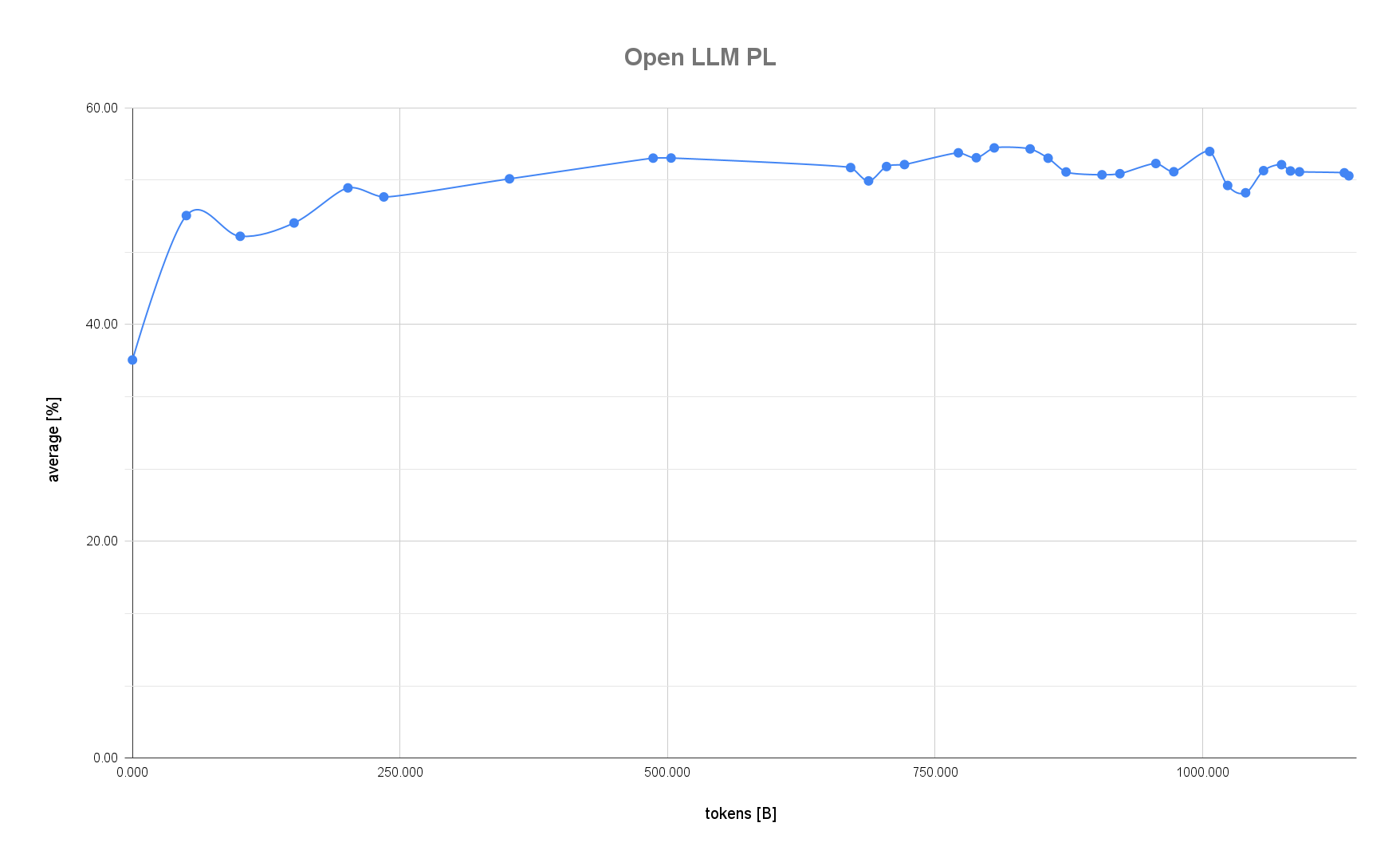}
\caption{Average score of the Bielik 11B v3 model checkpoints during pretraining in the Open PL LLM benchmark}
\label{fig:pretraining-OpenLLM_PL}
\end{figure}

To guarantee the desired quality throughout the pre-training execution, we diligently tracked the model's performance evolution on the Open LLM PL \cite{open-pl-llm-leaderboard} and Open LLM EN \cite{open-llm-leaderboard} benchmark suites. Maintaining performance on both of these benchmarks is crucial for preventing model degradation. Model checkpoints were captured at predefined intervals during training and subsequently subjected to performance evaluation, as shown in Figure \ref{fig:pretraining-OpenLLM_PL}. Our observations indicate that a decline in benchmark scores over the training period typically signals problems within the dataset and seldom proves reversible. Furthermore, a significant deterioration in either benchmark usually serves as an early indicator of wider model quality issues.

The performance assessment, utilizing the specified benchmarks, was executed concurrently with the core training on an isolated cluster of compute nodes to avoid resource contention. This strategy facilitated the prompt identification of training anomalies, eliminating the need to wait for full epoch completion and thus yielding considerable efficiencies in computational resources.

\section{Post-training}

Following the initial pre-training phase, we executed a rigorous post-training pipeline designed to bolster the model's capabilities in logical reasoning, mathematics, code generation, and complex instruction following. This stage followed a three-tier training protocol, mirroring the methodology developed for the smaller Bielik v3 model variants \cite{ociepa2025bielikv3smalltechnical}.

\subsection{Post-training Data} \label{Post-training-Data}

To address the scarcity of high-quality instruction-tuning data for the Polish language, we developed a proprietary dataset that undergoes continuous expansion and refinement by human annotators. This corpus is manually curated through the formulation of diverse instructions and multi-turn dialogues, ensuring the content is culturally relevant and linguistically precise for Polish and other European languages.

Furthermore, we utilized documents from the pre-training phase to synthesize instructions and dialogues via controlled prompting techniques. This approach ensures a broad representation of task types, inquiry styles, and domain-specific nuances. Throughout this synthetic generation process, we maintained a strict focus on factual consistency, linguistic naturalness, and structural clarity to match the standards of high-quality, human-authored data.

\subsection{Supervised Fine-Tuning}

The Supervised Fine-Tuning (SFT) stage is designed to achieve several critical objectives beyond basic task execution. It enables the model to identify and adapt to various registers and levels of formality, maintain long-range coherence in multi-turn conversations, and respond accurately to system-level prompts. Additionally, the model is trained to support structured outputs and function calling (tool use).

To optimize the SFT process, we implemented the following technical strategies:
\begin{itemize}
    \item \textbf{Constant Learning Rate and Weight Decay:} Employed to maintain stable optimization throughout the epoch cycles.
    \item \textbf{Instruction Masking:} We utilized a selective masking strategy where the loss function is calculated only on the model's responses. By masking user instructions and control tokens \cite{shi2024instructiontuninglossinstructions}, we ensured the gradient updates remained focused on content generation rather than predicting the prompt structure.
    \item \textbf{Sample Packing and FlexAttention:} To maximize computational throughput, we employed sample packing to concatenate multiple sequences up to the maximum context window, significantly outperforming traditional padding methods \cite{wang2024packinganalysispackingappropriate}. We integrated \texttt{FlexAttention} \cite{dong2024flexattentionprogrammingmodel} to generate attention masks dynamically, preventing cross-contamination between independent dialogues within a single packed sequence.
    \item \textbf{Selective Gradient Checkpointing:} This was applied to specific layers to achieve an optimal balance between memory conservation and processing speed.
    \item \textbf{Checkpoint Merging:} We performed weight averaging (merging) across various SFT training stages. This resulted in a final model that demonstrated superior evaluation performance compared to any individual intermediate checkpoint.
\end{itemize}

\subsubsection{Supervised Fine-Tuning Hyperparameters}

Optimization was performed using AdamW ($\beta_1 = 0.9$, $\beta_2 = 0.95$) with a weight decay of $0.05$. We maintained a constant learning rate of $5 \times 10^{-6}$ following a $100$-iteration linear warmup. The model was trained for $3$ epochs on $20$ million instructions and dialogues using a global batch size of $64$ and a maximum sequence length of $32,768$ tokens. Training utilized \texttt{bfloat16} mixed precision with a gradient clipping threshold of $1.0$.

\subsection{Preference Learning}

To align the Bielik 11B v3 model with human expectations, we employed \textbf{Direct Preference Optimization (DPO)} and its variant, \textbf{DPO-Positive (DPO-P)}, which proved highly effective in previous iterations of the Bielik series. 

\subsubsection{Preference Dataset}

Compared to previous versions of the Bielik model, we introduced a series of substantial upgrades to our Polish preference instruction corpus and the associated response generation framework. We expanded the dataset to over $114,000$ examples, prioritizing a more sophisticated distribution of tasks.

Key enhancements in the dataset composition include:
\begin{itemize}
    \item \textbf{Domain Expansion:} We integrated a higher volume of instructions focused on logical reasoning and mathematics.
    \item \textbf{Advanced Capabilities:} The alignment categories were broadened to explicitly include function calling and tool-use scenarios.
    \item \textbf{Interaction Realism:} To better simulate authentic user behavior, we incorporated an extensive collection of multi-turn conversational dialogues.
\end{itemize}

For the generation of contrastive response pairs (preferred vs.\ rejected), we diversified our model ensemble. In addition to our legacy pipeline, we employed several state-of-the-art models, most notably DeepSeek-V3-0324 and Bielik 11B v2.6.

Despite the increase in scale, our core philosophy regarding data integrity remains unchanged. We continue to employ a hybrid curation strategy that balances manual authorship and perturbation-based augmentation with automated safeguards. This strategy includes rigorous deduplication protocols, quality assessment via reward metamodels, and thorough manual audits. Collectively, these refinements ensure that the dataset provides a high-fidelity signal for downstream alignment and preference learning.

\subsubsection{DPO-P Hyperparameters}

The DPO-P phase utilized the AdamW optimizer with a constant learning rate of $5 \times 10^{-7}$ and $50$ warmup iterations. Training was conducted over $3$ epochs on the $114,000$ samples with a context window of $32,768$ tokens and a global batch size of $64$. Consistent with the SFT phase, we employed \texttt{bfloat16} mixed precision and a gradient clipping norm of $1.0$.

\subsection{Reinforcement Learning}

In the final stage of our alignment pipeline, we utilized Reinforcement Learning (RL) to further sharpen the model's analytical capabilities. We implemented Group Relative Policy Optimization (GRPO) \cite{shao2024deepseekmath} alongside its variant, Dr. GRPO \cite{liu2025understandingr1zeroliketrainingcritical}. The latter was specifically selected to improve token efficiency by mitigating the common issue where models artificially inflate response length to maximize rewards.

The RL training was performed using the Volcano Engine Reinforcement Learning (VERL) framework \cite{sheng2024hybridflow}, which provided a scalable and modular environment for our workflows. The training corpus consisted of 143k curated problems across logic, STEM, mathematics, and tool-use domains. Each sample was selected based on the availability of Reinforcement Learning from Verifiable Rewards (RLVR), ensuring that every problem had a definitive, checkable solution.

To facilitate accurate reward modeling, we developed specialized verifiers tailored to the task type:
\begin{itemize}
    \item \textbf{Mathematics:} Rewards were granted based on the extraction of the correct final answer, which was required to be enclosed within a \LaTeX{} \texttt{\textbackslash boxed\{...\}} command.
    \item \textbf{STEM:} For multiple-choice questions, the verifier assessed whether the model selected the single correct option from the available choices.
    \item \textbf{Tool Use:} The reward system validated whether the model invoked the appropriate function calls with the correct and expected arguments.
\end{itemize}

The integration of Reinforcement Learning significantly boosted the model's proficiency in mathematical tasks and broadly enhanced its general reasoning and problem-solving faculties. These results indicate that this training phase effectively generalizes across diverse domains, reinforcing the model's ability to follow complex logical paths beyond simple instruction following.

\subsubsection{Reinforcement Learning Hyperparameters}

Our optimization configuration utilized a learning rate of $1 \times 10^{-6}$ and a global batch size of 128 (distributed as local batches of 16 per GPU). To maintain stability during policy updates and prevent catastrophic forgetting, we applied KL divergence regularization with a coefficient of 0.001, employing a low-variance KL loss formulation.

\section{Evaluation}

To comprehensively assess the capabilities of Bielik v3 models, we conducted extensive evaluations across multiple benchmarks covering diverse aspects of language understanding, generation, and reasoning. Our evaluation strategy encompasses both Polish-specific and multilingual benchmarks to demonstrate the models' proficiency in handling various linguistic tasks.

The models were evaluated on the following benchmarks:

\begin{itemize}
    \item \href{https://huggingface.co/spaces/speakleash/open_pl_llm_leaderboard}{Open PL LLM Leaderboard} \textbf{(Polish)}
    \item \href{https://huggingface.co/spaces/speakleash/polish_eq-bench}{Polish EQ-Bench} \textbf{(Polish)}
    \item \href{https://huggingface.co/spaces/speakleash/cptu_bench}{CPTUB Leaderboard} \textbf{(Polish)}
    \item \href{https://huggingface.co/spaces/speakleash/polish_medical_leaderboard}{Polish Medical Leaderboard} \textbf{(Polish)}
    \item \href{https://huggingface.co/spaces/sdadas/plcc}{Polish Linguistic and Cultural Competency Benchmark (PLCC)} \textbf{(Polish)}
    \item \href{https://huggingface.co/spaces/open-llm-leaderboard-old/open_llm_leaderboard}{Open LLM Leaderboard}
    \item \href{https://huggingface.co/spaces/speakleash/include-base-european-leaderboard}{include-base-44}
    \item \href{https://huggingface.co/spaces/speakleash/belebele-european-leaderboard}{belebele}
    \item \href{https://huggingface.co/spaces/speakleash/leaderboard-flores}{flores}
\end{itemize}

\subsection{Open PL LLM Leaderboard} \label{Open-PL-LLM-Leaderboard}
The Open PL LLM Leaderboard \cite{open-pl-llm-leaderboard, ociepa2024bielik7bv01polish} provides a comprehensive assessment of language models across a diverse range of Polish NLP tasks. Built upon the foundation of Open LLM Leaderboard v1 \cite{open-llm-leaderboard-v1}, this benchmark evaluates core language understanding capabilities including sentiment classification, named entity recognition, topic categorization, reading comprehension, and question answering. The evaluation framework employs the lm-evaluation-harness toolkit \cite{eval-harness} and primarily focuses on discrete task performance rather than conversational interaction abilities.

\paragraph{Tasks:}
\begin{itemize}
    \item \textbf{polemo2:} Sentiment analysis of online consumer reviews across four domains (medicine, hotels, products, university) with four-class labeling (positive, negative, neutral, ambiguous) \cite{kocon-etal-2019-multi}; metric: accuracy.
    \item \textbf{klej-ner:} Named entity recognition in sentences containing single-type entities, classifying into six categories (no entity, place, person, organization, time, geographical name) \cite{rybak-etal-2020-klej}; metric: accuracy.
    \item \textbf{8tags:} Topic classification of social media headlines into eight categories (film, history, food, medicine, motorization, work, sport, technology) \cite{dadas-etal-2020-evaluation}; metric: accuracy.
    \item \textbf{belebele:} Machine reading comprehension for question answering \cite{bandarkar-etal-2024-belebele}; metric: accuracy.
    \item \textbf{dyk:} Question answering based on human-annotated pairs from Wikipedia's "Did You Know" section \cite{marcinczuk2013open}; metric: binary F1.
    \item \textbf{ppc:} Text similarity assessment using manually labeled sentence pairs (exact paraphrases, close paraphrases, non-paraphrases) \cite{9945218}; metric: accuracy.
    \item \textbf{psc:} Summarization of news articles \cite{ogro:kop:14:lrec}; metric: binary F1.
    \item \textbf{cbd:} Text classification for cyberbullying and hate-speech detection \cite{ptaszynski2023expert}; metric: macro F1.
    \item \textbf{polqa:} Open-domain question answering from the "Jeden z dziesi\k{e}ciu" TV show, with and without context (abstractive QA/RAG) \cite{rybak-etal-2024-polqa-polish}; metric: accuracy, levenshtein.
    \item \textbf{poquad:} Context-based extractive question answering (QA/RAG) \cite{tuora2023poquad}; metric: levenshtein.
    \item \textbf{eqbench:} emotional intelligence benchmark \cite{paech2024eqbenchemotionalintelligencebenchmark}; metric: custom.
\end{itemize}

The majority of benchmark tasks employ a multiple-choice format where models select from predefined answer options. Two distinct evaluation methodologies are applied:
\begin{itemize}
    \item \textbf{Loglikelihood:} Models select the option with the highest token probability from the available choices (e.g., A, B, C, D). This approach is particularly well-suited for evaluating base models without instruction tuning.
    \item \textbf{Generate:} Models produce free-form text responses, testing their generation capabilities.
\end{itemize}

Each task undergoes evaluation in both 0-shot (no examples provided) and 5-shot (five examples given) configurations, with final scores representing the average performance across all tasks, normalized against established baselines. Tables~\ref{tab:open-pl-llm-base} and~\ref{tab:open-pl-llm-instruct} present the results for base and instruction-tuned models, respectively.

\begin{table*}[t]
  \centering
  \small
  \begin{tabular}{lrr}
  \toprule
  \textbf{Model} & \textbf{Parameters (B)} & \textbf{Average} \\
  \midrule
  Qwen2.5-72B & 72.7 & 67.38 \\
  Qwen2.5-32B & 32.8 & 66.73 \\
  Qwen-72B & 72.7 & 66.02 \\
  Qwen2.5-14B & 14.8 & 62.71 \\
  Meta-Llama-3-70B & 70.6 & 62.07 \\
  Qwen1.5-72B & 72.7 & 61.11 \\
  Meta-Llama-3.1-70B & 70.6 & 60.87 \\
  Mixtral-8x22B-v0.1 & 141.0 & 60.75 \\
  Mistral-Small-24B-Base-2501 & 24.0 & 59.90 \\
  Qwen1.5-32B & 32.8 & 58.71 \\
  \underline{Bielik-11B-v2} & \underline{11.2} & \underline{58.14} \\
  Llama-4-Scout-17B-16E & 109.0 & 57.77 \\
  \textbf{Bielik-11B-v3-Base-20250730} & \textbf{11.2} & \textbf{55.16} \\
  Qwen2.5-7B & 7.0 & 53.35 \\
  EuroLLM-9B & 9.2 & 50.03 \\
  Qwen-7B & 7.0 & 49.39 \\
  SOLAR-10.7B-v1.0 & 10.7 & 47.54 \\
  Mistral-Nemo-Base-2407 & 12.2 & 47.28 \\
  internlm2-20b & 20.0 & 47.15 \\
  \underline{Bielik-4.5B-v3} & \underline{4.8} & \underline{45.47} \\
  Meta-Llama-3.1-8B & 8.0 & 43.77 \\
  Meta-Llama-3-8B & 8.0 & 43.30 \\
  Qwen1.5-72B & 72.3 & 39.51 \\
  Mistral-7B-v0.3 & 7.0 & 38.88 \\
  Mistral-7B-v0.2 & 7.0 & 38.81 \\
  Qwen1.5-7B & 7.0 & 37.92 \\
  \underline{Bielik-7B-v0.1} & \underline{7.2} & \underline{34.34} \\
  Qra-13b & 13.0 & 33.90 \\
  Llama-3.2-3B & 3.0 & 31.89 \\
  Qra-7b & 7.0 & 16.60 \\
  \bottomrule
  \end{tabular}
  \caption{Open PL LLM Leaderboard results for base models (5-shot evaluation)}
  \label{tab:open-pl-llm-base}
  \end{table*}

  \begin{table*}[t]
    \centering
    \small
    \begin{tabular}{lrr}
    \toprule
    \textbf{Model} & \textbf{Parameters (B)} & \textbf{Average} \\
    \midrule
    Mistral-Large-Instruct-2411 & 123.0 & 69.84 \\
    Meta-Llama-3.1-405B-Instruct-FP8 & 405.0 & 69.44 \\
    Mistral-Large-Instruct-2407 & 123.0 & 69.11 \\
    Qwen2.5-72B-Instruct & 72.7 & 67.92 \\
    QwQ-32B-Preview & 32.8 & 67.01 \\
    Llama-3.3-70B-Instruct & 70.6 & 66.40 \\
    \textbf{Bielik-11B-v3.0-Instruct} & \textbf{11.2} & \textbf{65.93} \\
    Qwen2-72B-Instruct & 72.7 & 65.87 \\
    \underline{Bielik-11B-v2.3-Instruct} & \underline{11.2} & \underline{65.71} \\
    \underline{Bielik-11B-v2.2-Instruct} & \underline{11.2} & \underline{65.57} \\
    Meta-Llama-3.1-70B-Instruct & 70.6 & 65.49 \\
    \underline{Bielik-11B-v2.1-Instruct} & \underline{11.2} & \underline{65.45} \\
    Mixtral-8x22B-Instruct-v0.1 & 141.0 & 65.23 \\
    \underline{Bielik-11B-v2.0-Instruct} & \underline{11.2} & \underline{64.98} \\
    Meta-Llama-3-70B-Instruct & 70.6 & 64.45 \\
    \underline{Bielik-11B-v2.6-Instruct} & \underline{11.2} & \underline{64.26} \\
    Qwen3-32B & 32.8 & 64.24 \\
    Llama-4-Scout-17B-16E-Instruct & 109.0 & 64.21 \\
    \underline{Bielik-11B-v2.5-Instruct} & \underline{11.2} & \underline{63.95} \\
    Mistral-Small-24B-Instruct-2501 & 24.0 & 62.97 \\
    phi-4 & 14.7 & 62.57 \\
    Qwen3-14B & 14.8 & 62.24 \\
    gemma-3-12b-it & 12.0 & 62.20 \\
    Mistral-Small-Instruct-2409 & 22.2 & 61.41 \\
    Qwen2.5-32B-Instruct & 32.8 & 61.21 \\
    Qwen2.5-14B-Instruct & 14.8 & 59.91 \\
    aya-23-35B & 35.0 & 56.37 \\
    \underline{Bielik-4.5B-v3.0-Instruct} & \underline{4.8} & \underline{56.13} \\
    gemma-3-27b-it & 27.0 & 55.92 \\
    Qwen3-8B & 8.2 & 55.78 \\
    Qwen3-4B & 4.0 & 55.49 \\
    Mistral-Nemo-Instruct-2407 & 12.2 & 55.27 \\
    EuroLLM-22B-Instruct-Preview & 22.0 & 55.17 \\
    Qwen2.5-7B-Instruct & 7.6 & 54.93 \\
    EuroLLM-9B-Instruct & 9.0 & 50.07 \\
    GaMS-9B-Instruct & 9.0 & 48.78 \\
    Mistral-7B-Instruct-v0.3 & 7.2 & 47.74 \\
    Apertus-8B-Instruct-2509 & 8.0 & 47.27 \\
    Mistral-7B-Instruct-v0.2 & 7.2 & 45.95 \\
    \underline{Bielik-7B-Instruct-v0.1} & \underline{7.2} & \underline{44.70} \\
    gemma-2-9b-it & 9.0 & 42.12 \\  
    Qwen2.5-3B-Instruct & 3.0 & 41.23 \\
    Mistral-7B-Instruct-v0.1 &	7.0 &	33.11 \\
    Qwen2.5-1.5B-Instruct & 1.5 & 31.89 \\
    \bottomrule
    \end{tabular}
    \caption{Open PL LLM Leaderboard results for instruction-tuned models (5-shot evaluation)}
    \label{tab:open-pl-llm-instruct}
    \end{table*}

As shown in Table~\ref{tab:open-pl-llm-base}, the Bielik-11B-v3-Base-20250730 model achieves a score of 55.16, demonstrating competitive performance for a base model of its size. The instruction-tuned variant, Bielik-11B-v3.0-Instruct, significantly outperforms its base counterpart with a score of 65.93 (Table~\ref{tab:open-pl-llm-instruct}), ranking among the top models and surpassing several much larger models including Meta-Llama-3.1-70B-Instruct and Mixtral-8x22B-Instruct-v0.1.

    \subsection{Polish EQ-Bench}

    The Polish Emotional Intelligence Benchmark represents a culturally adapted Polish adaptation of the EQ-Bench framework \cite{paech2024eqbenchemotionalintelligencebenchmark}. This benchmark assesses language models' ability to recognize, interpret, and reason about emotional states and interpersonal dynamics. The evaluation encompasses multiple facets of emotional intelligence, including emotion recognition in context, understanding of emotional implications, and sensitivity to nuanced affective states in conversational scenarios. Results are presented in Table~\ref{tab:pl-eq-bench}.
    
    \begin{table*}[t]
    \centering
    \small
    \begin{tabular}{lrc}
    \toprule
    \textbf{Model} & \textbf{Parameters (B)} & \textbf{Score} \\
    \midrule
    Mistral-Large-Instruct-2407$^{\dagger}$ & 123.0 & 78.07 \\
    Mistral-Large-Instruct-2411$^{\dagger}$ & 123.0 & 77.29 \\
    Meta-Llama-3.1-405B-Instruct-FP8 & 405.0 & 77.23 \\
    gpt-4o-2024-08-06 & Unknown & 75.15 \\
    gpt-4-turbo-2024-04-09 & Unknown & 74.59 \\
    \underline{Bielik-11B-v2.6-Instruct} & \underline{11.2} & \underline{73.8} \\
    Mistral-Small-Instruct-2409 & 22.2 & 72.85 \\
    Llama-PLLuM-70B-chat & 70.6 & 72.56 \\
    Meta-Llama-3.1-70B-Instruct & 70.6 & 72.53 \\
    DeepSeek-V3-0324 & 685.0 & 73.46 \\
    \underline{Bielik-11B-v2.5-Instruct} & \underline{11.2} & \underline{72.00} \\
    Qwen2-72B-Instruct & 72.7 & 71.23 \\
    Meta-Llama-3-70B-Instruct & 70.6 & 71.21 \\
    \textbf{Bielik-11B-v3.0-Instruct} & \textbf{11.2} & \textbf{71.20} \\
    gpt-4o-mini-2024-07-18 & Unknown & 71.15 \\
    Qwen2.5-32B-Instruct & 32.8 & 71.15 \\
    \underline{Bielik-11B-v2.3-Instruct} & \underline{11.2} & \underline{70.86} \\
    Llama-3.3-70B-Instruct & 70.6 & 70.73 \\
    Llama-PLLuM-70B-instruct & 70.6 & 69.99 \\
    WizardLM-2-8x22B & 141.0 & 69.56 \\
    Qwen2.5-14B-Instruct & 14.8 & 69.17 \\
    \underline{Bielik-11B-v2.2-Instruct} & \underline{11.2} & \underline{69.05} \\
    \underline{Bielik-11B-v2.0-Instruct} & \underline{11.2} & \underline{68.24} \\
    glm-4-9b-chat & 9.0 & 61.79 \\
    Mistral-Nemo-Instruct-2407 & 12.2 & 61.76 \\
    \underline{Bielik-11B-v2.1-Instruct} & \underline{11.2} & \underline{60.07} \\
    pllum-12b-nc-chat-250715 & 12.2 & 55.20 \\
    EuroLLM-9B-Instruct & 9.2 & 54.10 \\
    \underline{Bielik-4.5B-v3.0-Instruct} & \underline{4.8} & \underline{53.58} \\
    PLLuM-12B-chat & 12.2 & 52.26 \\
    PLLuM-8x7B-nc-chat$^{\dagger}$ & 46.7 & 47.29 \\
    Llama-PLLuM-8B-chat & 8.0 & 46.20 \\
    PLLuM-8x7B-chat & 46.7 & 45.22 \\
    PLLuM-12B-nc-chat$^{\dagger}$ & 12.2 & 35.41 \\
    \bottomrule
    \multicolumn{3}{l}{$^{\dagger}$Models with a non-commercial license.} \\
    \end{tabular}
    \caption{Polish EQ-Bench results for various models.}
    \label{tab:pl-eq-bench}
    \end{table*}

Bielik-11B-v3.0-Instruct achieves a score of 71.20 on the Polish EQ-Bench (Table~\ref{tab:pl-eq-bench}), demonstrating strong emotional intelligence capabilities. While this represents a slight decrease compared to the previous version Bielik-11B-v2.6-Instruct (73.8), the v3.0 model maintains competitive performance, placing it among models with substantially larger parameter counts such as Llama-3.3-70B-Instruct (70.73) and Qwen2.5-32B-Instruct (71.15).

    \subsection{Complex Polish Text Understanding Benchmark (CPTUB)}

    CPTUB \cite{cptub-leaderboard} presents a sophisticated evaluation framework targeting advanced comprehension capabilities in Polish language processing. In contrast to conventional benchmarks that primarily test literal interpretation, CPTUB probes deeper cognitive abilities including inference from context, pragmatic understanding, and reasoning under ambiguity. The benchmark structure encompasses two primary evaluation dimensions:
    
    \begin{itemize}
        \item \textbf{Implicatures}: This component measures models' competence in decoding non-literal meanings and contextual implications. It examines understanding of figurative language, ironic expressions, and idiomatic constructions through three distinct evaluation categories:
        \begin{itemize}
            \item \textbf{Sentiment}: Assessing the ability to discern emotional valence that diverges from surface-level lexical content
            \item \textbf{Language understanding}: Testing comprehension of communicative intent and pragmatic meaning
            \item \textbf{Phraseology}: Evaluating knowledge of conventionalized multi-word expressions where compositional semantics fail
        \end{itemize}
        \item \textbf{Tricky Questions}: This section challenges models with adversarially constructed queries featuring logical paradoxes, semantic ill-formedness, contradictions, absurdist premises, and humorous misdirection. It specifically probes reasoning robustness and the model's tendency to produce plausible-sounding but incorrect responses when confronted with problematic inputs.
    \end{itemize}
    
    Table~\ref{tab:cptub} presents the comprehensive results across all CPTUB evaluation categories.
    
    \begin{landscape}
    \begin{table*}[t]
    \centering
    \small
    \begin{tabular}{lrccccccc}
    \toprule
    \textbf{Model} & \textbf{Params (B)} & \textbf{Overall} & \textbf{Implicatures} & \textbf{Senti-} & \textbf{Language} & \textbf{Phrase-} & \textbf{Tricky} \\
    & & \textbf{Average} & \textbf{Average} & \textbf{ment} & \textbf{Understanding} & \textbf{ology} & \textbf{Questions} \\
    \midrule
    gemini-2.0-flash-001 & 4.29 & 4.39 & 4.52 & 4.32 & 4.34 & 3.99 \\
    DeepSeek-R1 & 685.0 & 4.14 & 4.14 & 4.49 & 4.35 & 3.60 & 4.12 \\
    gemini-2.0-flash-lite-001 & 4.09 & 4.17 & 4.23 & 4.05 & 4.24 & 3.85 \\
    DeepSeek-V3-0324 & 685.0 & 4.03 & 4.03 & 4.36 & 4.20 & 3.54 & 4.02 \\
    Mistral-Large-Instruct-2411$^{\dagger}$ & 123.0 & 4.00 & 4.10 & 4.33 & 3.98 & 3.99 & 3.72 \\
    Qwen2.5-72B-Instruct & 72.7 & 3.95 & 3.99 & 4.08 & 3.97 & 3.93 & 3.81 \\
    Mistral-Large-Instruct-2407$^{\dagger}$ & 123.0 & 3.93 & 4.03 & 4.23 & 4.00 & 3.86 & 3.65 \\
    Llama-4-Maverick-17B-128E-Instruct & 402.0 & 3.93 & 3.99 & 4.39 & 4.11 & 3.48 & 3.76 \\
    gemma-3-27b-it & 27.4 & 3.81 & 3.90 & 3.88 & 3.79 & 4.03 & 3.53 \\
    Meta-Llama-3-70B-Instruct & 70.6 & 3.78 & 3.81 & 4.13 & 3.82 & 3.47 & 3.71 \\
    Qwen2.5-32B-Instruct & 32.8 & 3.75 & 3.80 & 3.81 & 3.57 & 4.04 & 3.59 \\
    Llama-4-Scout-17B-16E-Instruct & 109.0 & 3.75 & 3.94 & 4.10 & 3.81 & 3.90 & 3.19 \\
    \textbf{Bielik-11B-v3.0-Instruct} & \textbf{11.2} & \textbf{3.73} & \textbf{3.92} & \textbf{3.88} & \textbf{3.91} & \textbf{3.96} & \textbf{3.19} \\
    Mistral-Small-24B-Instruct-2501 & 23.6 & 3.71 & 3.80 & 3.91 & 3.60 & 3.88 & 3.45 \\
    pllum-12b-nc-chat-250715$^{\dagger}$ & 12.2 & 3.67 & 3.92 & 4.36 & 3.96 & 3.46 & 2.90 \\
    \underline{Bielik-11B-v2.6-Instruct} & \underline{11.2} & \underline{3.64} & \underline{3.82} & \underline{4.10} & \underline{3.94} & \underline{3.41} & \underline{3.10} \\
    Mixtral-8x22B-Instruct-v0.1 & 141.0 & 3.56 & 3.67 & 3.78 & 3.68 & 3.55 & 3.24 \\
    Qwen2.5-14B-Instruct & 14.8 & 3.55 & 3.62 & 3.91 & 3.57 & 3.37 & 3.34 \\
    Llama-PLLuM-70B-chat & 70.6 & 3.53 & 3.63 & 3.94 & 3.61 & 3.35 & 3.21 \\
    \underline{Bielik-4.5B-v3.0-Instruct} & \underline{4.8} & \underline{3.38} & \underline{3.68} & \underline{3.76} & \underline{3.61} & \underline{3.67} & \underline{2.46} \\
    phi-4 & 14.7 & 3.30 & 3.50 & 3.72 & 3.54 & 3.24 & 2.72 \\
    PLLuM-12B-chat & 12.2 & 3.14 & 3.32 & 3.32 & 3.21 & 3.43 & 2.59 \\
    PLLuM-8x7B-nc-instruct$^{\dagger}$ & 46.7 & 3.11 & 3.56 & 3.88 & 3.59 & 3.22 & 1.76 \\
    Qwen2.5-7B-Instruct & 7.62 & 3.07 & 3.23 & 3.56 & 3.03 & 3.10 & 2.58 \\
    EuroLLM-9B-Instruct & 9.0 & 3.15 & 3.28 & 3.37 & 3.30 & 3.17 & 2.75 \\
    PLLuM-8x7B-nc-chat$^{\dagger}$ & 46.7 & 3.03 & 3.44 & 3.76 & 3.48 & 3.08 & 1.80 \\
    Meta-Llama-3.1-8B-Instruct & 8.0 & 3.01 & 3.31 & 3.97 & 3.38 & 2.58 & 2.11 \\
    PLLuM-8x7B-chat & 46.7 & 3.01 & 3.41 & 3.44 & 3.45 & 3.35 & 1.78 \\
    Meta-Llama-3-8B-Instruct & 8.0 & 3.00 & 3.17 & 3.33 & 3.15 & 3.04 & 2.48 \\
    Llama-PLLuM-8B-chat & 8.0 & 2.92 & 3.14 & 3.13 & 2.93 & 3.36 & 2.25 \\
    \underline{Bielik-7B-Instruct-v0.1} & \underline{7.2} & \underline{2.88} & \underline{3.13} & \underline{3.59} & \underline{3.48} & \underline{2.32} & \underline{2.16} \\
    \bottomrule
    \multicolumn{8}{l}{$^{\dagger}$Models with a non-commercial license.} \\
    \end{tabular}
    \caption{Complex Polish Text Understanding Benchmark (CPTUB) results across different evaluation categories}
    \label{tab:cptub}
    \end{table*}
\end{landscape}

On CPTUB (Table~\ref{tab:cptub}), Bielik-11B-v3.0-Instruct achieves an overall average of 3.73, ranking competitively among models evaluated. The model performs particularly well on implicature understanding with an average of 3.92, demonstrating strong capabilities in language understanding (3.91), sentiment analysis (3.88), and phraseology (3.96). The tricky questions component yields a score of 3.19, reflecting the challenging nature of these adversarial queries. This performance places Bielik-11B-v3.0-Instruct ahead of several larger models including Qwen2.5-14B-Instruct and Mixtral-8x22B-Instruct-v0.1, while approaching the performance of frontier models with significantly higher parameter counts.

    \subsection{Polish Medical Leaderboard}

    The Polish Medical Leaderboard provides a domain-specific assessment of language models using authentic questions from the Polish State Specialization Examination (Pa\'{n}stwowy Egzamin Specjalizacyjny, PES) spanning 2018-2022. This benchmark measures both medical domain knowledge and clinical reasoning abilities within the Polish healthcare context. The evaluation employs the speakleash/PES-2018-2022 dataset, derived from amu-cai/PES-2018-2022 \cite{pokrywka2024gpt4passes}, and tests models' capacity to apply medical knowledge in scenarios similar to those encountered by medical professionals seeking board certification. Results are shown in Table~\ref{tab:medical-leaderboard}.
    
    \begin{table*}[t]
    \centering
    \small
    \begin{tabular}{lrc}
    \toprule
    \textbf{Model} & \textbf{Parameters (B)} & \textbf{Average (\%)} \\
    \midrule
    Meta-Llama-3.1-405B-Instruct-FP8 & 405.0 & 69.20 \\
    Mistral-Large-Instruct-2407$^{\dagger}$ & 123.0 & 64.28 \\
    Qwen2.5-72B-Instruct & 72.7 & 63.89 \\
    Meta-Llama-3.1-70B-Instruct & 70.6 & 61.75 \\
    Qwen2-72B-Instruct & 72.7 & 61.35 \\
    Meta-Llama-3-70B-Instruct & 70.6 & 57.51 \\
    Qwen2.5-32B & 32.8 & 55.69 \\
    Qwen2.5-32B-Instruct & 32.8 & 54.52 \\
    \textbf{Bielik-11B-v3.0-Instruct} & \textbf{11.2} & \textbf{50.21} \\
    Qwen2.5-14B-Instruct & 14.8 & 49.60 \\
    \textbf{Bielik-11B-v3-Base-20250730} & \textbf{11.2} & \textbf{45.86} \\
    \underline{Bielik-11B-v2.6-Instruct} & \underline{11.2} & \underline{44.88} \\
    \underline{Bielik-11B-v2.5-Instruct} & \underline{11.2} & \underline{44.85} \\
    GLM-4-9b-chat & 9.0 & 44.54 \\
    Mistral-Small-Instruct-2409 & 22.2 & 43.60 \\
    \underline{Bielik-4.5B-v3.0-Instruct} & \underline{4.8} & \underline{43.55} \\
    \underline{Bielik-11B-v2.3-Instruct} & \underline{11.2} & \underline{43.26} \\
    \underline{Bielik-11B-v2.1-Instruct} & \underline{11.2} & \underline{43.16} \\
    \underline{Bielik-11B-v2.2-Instruct} & \underline{11.2} & \underline{43.05} \\
    Qwen2.5-7B-Instruct & 7.6 & 42.69 \\
    \underline{Bielik-11B-v2.0-Instruct} & \underline{11.2} & \underline{41.53} \\
    Meta-Llama-3.1-8B-Instruct & 8.0 & 40.60 \\
    Mistral-Nemo-Instruct-2407 & 12.2 & 40.36 \\
    \underline{Bielik-11B-v2} & \underline{11.2} & \underline{39.98} \\
    PLLuM-12B-nc-chat-250715$^{\dagger}$ & 12.2 & 38.53 \\
    PLLuM-12B-chat & 12.2 & 36.51 \\
    EuroLLM-9B-Instruct & 9.0 & 35.96 \\
    Mistral-7B-Instruct-v0.3 & 7.0 & 31.24 \\
    \underline{Bielik-7B-Instruct-v0.1} & \underline{7.2} & \underline{29.74} \\
    \bottomrule
    \multicolumn{3}{l}{$^{\dagger}$Models with a non-commercial license.} \\
    \end{tabular}
    \caption{Polish Medical Leaderboard results (5-shot setting) showing model performance on Polish Board Certification Examinations.}
    \label{tab:medical-leaderboard}
    \end{table*}

On the Polish Medical Leaderboard (Table~\ref{tab:medical-leaderboard}), Bielik-11B-v3.0-Instruct achieves 50.21\%, demonstrating substantial medical knowledge and clinical reasoning capabilities. This represents a significant improvement over the base model Bielik-11B-v3-Base-20250730 (45.86\%), highlighting the effectiveness of instruction tuning for specialized domain tasks. These results demonstrate Bielik's capability to handle domain-specific knowledge in the medical field when evaluated in Polish.

    \subsection{Polish Linguistic and Cultural Competency Benchmark (PLCC)}

    PLCC \cite{dadas2025evaluatingpolishlinguisticcultural} extends evaluation beyond linguistic proficiency to encompass cultural knowledge deeply embedded in Polish society. This benchmark features 600 manually curated questions spanning six thematic domains: historical events, geographical knowledge, cultural traditions, artistic and entertainment heritage, grammatical correctness, and lexical richness. 
    
    The evaluation probes understanding of culturally-specific references, including historical milestones, traditional customs, folklore narratives, literary works, and contemporary popular culture--all elements crucial for authentic communication and comprehension in Polish contexts. Question difficulty varies from widely recognized facts to regionally-specific knowledge, employing both multiple-choice and open-ended response formats that demand precise factual recall of dates, proper nouns, and conceptual relationships. Performance results are presented in Table~\ref{tab:plcc-scores}.
    
    \begin{table*}[t]
    \centering
    \small
    \begin{tabular}{lrc}
    \toprule
    \textbf{Model} & \textbf{Parameters (B)} & \textbf{Average Score (\%)} \\
    \midrule
    Gemini-3.0-Pro-Preview & Unknown & 95.83 \\
    Gemini-2.5-Pro-Exp-03-25	& Unknown &	89.50 \\
    Claude-3.7-Sonnet & Unknown & 81.50 \\
    GPT-5.1-2025-11-13 (default reasoning) & Unknown & 77.83 \\
    Gemini-2.0-Flash-Experimental & Unknown & 74.17 \\
    O4-Mini-2025-04-16 & Unknown & 72.83 \\
    \textbf{Bielik-11B-v3.0-Instruct} &  \textbf{11.2} & \textbf{71.83} \\
    Kimi-K2-Thinking &  1000 & 71.67 \\
    Claude-Sonnet-4.5 & Unknown & 71.00 \\
    DeepSeek-v3-0324	& 685.0 &	71.00 \\
    GLM-4.6 & 355 & 70.67 \\
    Grok-4-Fast & Unknown & 70.17 \\
    Gemini-Pro-1.5 & Unknown & 69.67 \\
    PLLuM-12B-nc-chat-250715$^{\dagger}$ & 12.2 & 69.67 \\
    DeepSeek-v3		& 685.0 & 69.17 \\
    PLLuM-8x7B-nc-chat$^{\dagger}$ & 46.7 & 68.17 \\
    \underline{Bielik-11B-v2.6-Instruct} & \underline{11.2} & \underline{65.50} \\
    GPT-4.1-mini-2025-04-14	& Unknown &	62.17 \\
    Llama-3.1-405B & 405.0 & 60.00 \\
    PLLuM-12B-nc-chat$^{\dagger}$ & 12.2 & 59.50 \\
    Llama-PLLuM-70B-chat & 70.6 & 58.50 \\
    Llama-4-Maverick & 402.0 & 58.17 \\
    Command-A-03-2025$^{\dagger}$ & 111.0 & 56.17 \\
    Mistral-Large-2407$^{\dagger}$ & 123.0 & 54.17 \\
    PLLuM-8x7B-chat & 46.7 & 54.17 \\
    WizardLM-2-8x22B & 141.0 & 51.50 \\
    Qwen-Max & Unknown & 50.83 \\
    Command-R-Plus-08-2024$^{\dagger}$ & Unknown & 50.17 \\
    Mixtral-8x22B & 141.0 & 49.83 \\
    Llama-3.3-70B & 70.6 & 48.83 \\
    Gemma-3-27B & 27.4 & 47.33 \\
    PLLuM-12B-chat & 12.2 & 47.00 \\
    \underline{Bielik-7B-Instruct-v0.1} & \underline{7.2} & \underline{46.67} \\
    Mistral-Small-3.1-24B-2503 & 24.0 & 43.33 \\
    Llama-3.0-70B & 70.0 & 43.00 \\
    Gemma-2-27B & 27.0 & 42.67 \\
    Llama-4-Scout & 109.0 & 41.50 \\
    \underline{Bielik-4.5B-v3.0-Instruct} & \underline{4.8} & \underline{41.47} \\
    EuroLLM-9B & 9.0 & 41.00 \\
    Qwen-2.5-72B & 72.7 & 39.17 \\
    Mistral-Small-24B-2501 & 24.0 & 39.00 \\
    Llama-PLLuM-8B-chat & 8.0 & 38.50 \\
    Qwen3-32B & 32.8 & 37.67 \\
    Mixtral-8x7B & 46.7 & 35.33 \\
    GPT-OSS-20b & 20 & 32.3 \\
    Qwen-2.5-32B & 32.8 & 30.50 \\
    Gemma-2-9B & 9.0 & 29.17 \\
    Phi-4 & 14.7 & 29.17 \\
    \underline{Bielik-1.5B-v3.0-Instruct} & \underline{1.6} & \underline{27.50} \\
    Qwen-2.5-14B & 14.8 & 26.67 \\
    Mistral-Nemo & 12.2 & 23.00 \\
    Mistral-7B-v0.3 & 7.2 & 21.83 \\
    \bottomrule
    \multicolumn{3}{l}{$^{\dagger}$Models with a non-commercial license.} \\
    \end{tabular}
    \caption{Polish Linguistic and Cultural Competency Benchmark (PLCC) results for open-source models. Closed proprietary models have been excluded from this comparison.}
    \label{tab:plcc-scores}
    \end{table*}

Bielik-11B-v3.0-Instruct achieves an exceptional score of 71.83\% on PLCC (Table~\ref{tab:plcc-scores}), demonstrating superior cultural and linguistic knowledge compared to other open-source models evaluated. This represents a substantial improvement over the previous Bielik-11B-v2.6-Instruct (65.50\%) and significantly surpasses much larger models including DeepSeek-V3-0324 (71.00\%, 685B parameters), Llama-3.1-Tulu-3-405B (63.83\%). These results underscore Bielik's particular strength in capturing Polish-specific cultural knowledge, a critical capability for applications requiring deep understanding of Polish context.

    \subsection{Open LLM Leaderboard}

    The Open LLM Leaderboard \cite{open-llm-leaderboard} serves as a comprehensive English-language evaluation suite, assessing models across diverse tasks including commonsense reasoning (ARC challenge, HellaSwag, WinoGrande), factual accuracy (TruthfulQA), broad knowledge (MMLU), and mathematical reasoning (GSM8K). This benchmark provides crucial insights into multilingual models' English language capabilities, which is particularly important for European models like Bielik that aim to balance strong native language performance with English proficiency. Tables~\ref{tab:open-llm-base} and~\ref{tab:open-llm-instruct} present results for base and instruction-tuned models, respectively.

    \begin{landscape}
    \begin{table*}[t]
    \centering
    \small
    \begin{tabular}{lccccccc}
    \toprule
    \textbf{Model} & \textbf{AVG} & \textbf{arc\_challenge} & \textbf{hellaswag} & \textbf{truthfulqa\_mc2} & \textbf{mmlu} & \textbf{winogrande} & \textbf{gsm8k} \\
    \midrule
    \textbf{Bielik-11B-v3} & \textbf{68.45} & \textbf{61.43}	& \textbf{81.38}	& \textbf{47.65}	& \textbf{67.55}	& \textbf{78.53}	& \textbf{74.15} \\
    Qwen1.5-14B & 66.70 & 56.57 & 81.08 & 52.06 & 69.36 & 73.48 & 67.63 \\
    \underline{Bielik-11B-v2} & \underline{65.87} & \underline{60.58} & \underline{79.84} & \underline{46.13} & \underline{63.06} & \underline{77.82} & \underline{67.78} \\
    Qwen-14B & 65.86 & 58.28 & 83.99 & 49.43 & 67.70 & 76.80 & 58.98 \\
    Meta-Llama-3-8B & 62.62 & 60.24 & 82.23 & 42.93 & 66.70 & 78.45 & 45.19 \\
    \underline{Bielik-4.5B-v3} & \underline{61.02} & \underline{51.19} & \underline{73.01} & \underline{45.63} & \underline{61.32} & \underline{71.35} & \underline{63.61} \\
    Mistral-7B-v0.1 & 60.97 & 59.98 & 83.31 & 42.15 & 64.16 & 78.37 & 37.83 \\
    Mistral-7B-v0.2 & 60.37 & 60.84 & 83.08 & 41.76 & 63.62 & 78.22 & 34.72 \\
    \underline{Bielik-7B-v0.1} & \underline{49.98} & \underline{45.22} & \underline{67.92} & \underline{47.16} & \underline{43.20} & \underline{66.85} & \underline{29.49} \\
    \bottomrule
    \end{tabular}
    \caption{Open LLM Leaderboard results for base models}
    \label{tab:open-llm-base}
    \end{table*}
    \end{landscape}
 
    \begin{landscape}
    \begin{table*}[t]
    \centering
    \small
    \begin{tabular}{lccccccc}
    \toprule
    \textbf{Model} & \textbf{AVG} & \textbf{arc\_challenge} & \textbf{hellaswag} & \textbf{truthfulqa\_mc2} & \textbf{mmlu} & \textbf{winogrande} & \textbf{gsm8k} \\
    \midrule
    SOLAR-10.7B-Instruct-v1.0 & 74.20 & 71.08 & 88.16 & 71.43 & 66.21 & 83.58 & 64.75 \\
    Phi-3-medium-4k-instruct & 73.45 & 67.32 & 85.76 & 57.71 & 77.83 & 72.69 & 79.38 \\
    \textbf{Bielik-11B-v3.0-Instruct} & \textbf{72.45} & \textbf{64.59} & \textbf{81.96} & \textbf{54.25} & \textbf{71.11} & \textbf{77.19} & \textbf{85.60} \\
    \underline{Bielik-11B-v2.5-Instruct} & \underline{71.42} & \underline{61.95} & \underline{80.71} & \underline{53.17} & \underline{67.44} & \underline{79.72}	& \underline{85.52} \\
    \underline{Bielik-11B-v2.6-Instruct} & \underline{71.10} & 62.54 & 80.56 & 53.43 & 67.53 & 78.77 & 83.78 \\
    Bielik-11B-v2.2-Instruct & 69.86 & 59.90 & 80.16 & 58.34 & 64.34 & 75.30 & 81.12 \\
    \underline{Bielik-11B-v2.3-Instruct} & \underline{69.82} & 59.30 & 80.11 & 57.42 & 64.57 & 76.24 & 81.27 \\
    Bielik-11B-v2.1-Instruct & 69.82 & 59.56 & 80.20 & 59.35 & 64.18 & 75.06 & 80.59 \\
    openchat-3.5-0106-gemma & 69.42 & 64.68 & 81.08 & 54.93 & 64.69 & 78.30 & 72.86 \\
    Bielik-11B-v2.0-Instruct & 68.04 & 58.62 & 78.65 & 54.65 & 63.71 & 76.32 & 76.27 \\
    Meta-Llama-3-8B-Instruct & 66.87 & 60.75 & 78.55 & 51.65 & 67.07 & 74.51 & 68.69 \\
    Mistral-7B-Instruct-v0.2 & 65.71 & 63.14 & 84.88 & 68.26 & 60.78 & 77.19 & 40.03 \\
    \underline{Bielik-4.5B-v3-Instruct} & \underline{64.89} & \underline{56.06} & \underline{73.90} & \underline{50.79} & \underline{63.66} & \underline{71.19} & \underline{73.69} \\
    gemma-7b & 64.29 & 61.09 & 82.47 & 44.91 & 66.03 & 78.45 & 52.77 \\
    Qwen1.5-32B-Chat & 62.95 & 66.04 & 85.49 & 66.95 & 74.99 & 77.19 & 7.05 \\
    Qwen1.5-14B-Chat & 62.27 & 58.70 & 82.27 & 60.36 & 68.57 & 73.09 & 30.63 \\
    Qwen1.5-7B-Chat & 55.15 & 55.89 & 78.56 & 53.54 & 61.65 & 67.72 & 13.57 \\
    Mistral-7B-Instruct-v0.1 & 54.96 & 54.52 & 75.63 & 56.28 & 55.38 & 73.72 & 14.25 \\
    \underline{Bielik-7B-Instruct-v0.1} & \underline{51.26} & 47.53 & 68.91 & 46.18 & 49.47 & 65.51 & 29.95 \\
    \bottomrule
    \end{tabular}
    \caption{Open LLM Leaderboard results for selected instruction-tuned models}
    \label{tab:open-llm-instruct}
    \end{table*}
\end{landscape}

On English language tasks (Tables~\ref{tab:open-llm-base} and~\ref{tab:open-llm-instruct}), Bielik models demonstrate solid cross-lingual capabilities. The base Bielik-11B-v2 reaches 65.87 average score. For instruction-tuned models, Bielik-11B-v3.0-Instruct scores 72.45 average, with particularly strong performance on GSM8K (85.60) and ARC challenge (64.59), indicating robust mathematical and reasoning capabilities.

\subsection{INCLUDE-base-44}

INCLUDE \cite{romanou2024include} is a comprehensive knowledge- and reasoning-centric benchmark designed to evaluate multilingual language models across 44 languages in realistic deployment scenarios. The benchmark comprises 22,637 four-option multiple-choice questions extracted from academic and professional examinations, covering 57 topics across diverse domains including STEM (Biology, Chemistry, Physics, Mathematics, Computer Science), Arts \& Humanities (History, Philosophy, Literature, Visual Arts, Law), Social Sciences (Sociology, Economics, Psychology), Health-oriented Education (Medicine), and professional certifications (driving licenses, medical licenses, professional certifications).

A distinguishing feature of INCLUDE is its emphasis on regional knowledge and cultural context. Questions are categorized as either "agnostic" (universally applicable) or "region implicit/explicit" (requiring cultural or geographical knowledge specific to particular regions). This design enables assessment of models' ability to handle not only universal knowledge but also culturally-specific content essential for deployment in diverse linguistic communities. For our evaluation, we focus on a subset of 20 European languages from the full benchmark to assess Bielik's performance across its target linguistic region. The benchmark evaluation is presented in Table~\ref{tab:include-base-44}.

\begin{table*}[t]
  \centering
  \small
  \begin{tabular}{lrrr}
  \toprule
  \textbf{Model} & \textbf{Params (B)} & \textbf{AVG} & \textbf{Polish} \\
  \midrule
  \textbf{Bielik-11B-v3-Instruct} & \textbf{11} & \textbf{64.8} & \textbf{69.0} \\
  Qwen2.5-14B-Instruct & 14 & 61.7 & 58.9 \\
  \textbf{Bielik-11B-v3} & \textbf{11} & \textbf{60.6} & \textbf{63.9} \\
  phi-4 & 15 & 58.8 & 49.6 \\
  Apertus-8B-Instruct-2509 & 8 & 57.9 & 49.6 \\
  Llama-3.1-8B-Instruct & 8 & 55.3 & 53.8 \\
  EuroLLM-9B-Instruct & 9 & 55.1 & 52.0 \\
  Qwen2.5-7B-Instruct & 7 & 54.4 & 52.2 \\
  Mistral-Nemo-Instruct-2407 & 12 & 53.2 & 48.4 \\
  \underline{Bielik-11B-v2.6-Instruct} & \underline{11} & \underline{51.5} & \underline{59.3} \\
  Mistral-Nemo-Base-2407 & 12 & 51.2 & 44.9 \\
  EuroLLM-9B & 9 & 49.2 & 45.6 \\
  aya-expanse-8b & 8 & 45.3 & 46.4 \\
  Mistral-7B-Instruct-v0.2 & 7 & 45.3 & 44.7 \\
  \underline{Bielik-11B-v2} & \underline{11} & \underline{44.8} & \underline{53.5} \\
  pllum-12b-nc-chat-250715 & 12 & 44.2 & 60.6 \\
  Mistral-7B-v0.2 & 7 & 41.8 & 37.2 \\
  pllum-12b-nc-base-250715 & 12 & 37.8 & 52.7 \\
  \underline{Bielik-4.5B-v3} & \underline{4.5} & \underline{35.9} & \underline{48.7} \\
  PLLuM-12B-base-250801 & 12 & 35.5 & 44.5 \\
  Llama-PLLuM-8B-base-250801 & 8 & 30.0 & 37.2 \\
  \bottomrule
  \end{tabular}
  \caption{INCLUDE-base-44 benchmark results showing average performance across European languages (20 language subset) and Polish-specific scores.}
  \label{tab:include-base-44}
\end{table*}

On INCLUDE-base-44 (Table~\ref{tab:include-base-44}), Bielik-11B-v3-Instruct achieves the highest scores among all evaluated models, with 64.8 overall average across European languages and 69.0 on Polish-specific tasks. This demonstrates superior balanced multilingual performance, surpassing Qwen2.5-14B-Instruct (61.7 average, 58.9 Polish) despite having fewer parameters. Notably, Bielik's Polish-specific score (69.0) substantially exceeds its multilingual average (64.8), reflecting the model's particular strength in its primary target language while maintaining robust cross-lingual capabilities. The base model Bielik-11B-v3 also shows strong performance (60.6 average, 63.9 Polish), outperforming several instruction-tuned models including Llama-3.1-8B-Instruct and EuroLLM-9B-Instruct. Compared to the previous version, Bielik-11B-v2 achieved 44.8 average with 53.5 on Polish tasks, showing significant improvement in v3.

\subsection{Belebele Reading Comprehension}

Belebele \cite{bandarkar-etal-2024-belebele} is a massively multilingual reading comprehension benchmark spanning 122 language variants. The benchmark consists of multiple-choice reading comprehension questions derived from the FLORES-200 dataset, where models must demonstrate understanding of short passages by correctly answering questions about their content. For our evaluation, we assess performance across 28 European language variants to evaluate Bielik's reading comprehension capabilities across its target linguistic region. Results are presented in Table~\ref{tab:belebele}.

\begin{table*}[t]
  \centering
  \small
  \begin{tabular}{lrr}
  \toprule
  \textbf{Model} & \textbf{Params (B)} & \textbf{Score} \\
  \midrule
  Qwen2.5-14B-Instruct & 14 & 85.91 \\
  \textbf{Bielik-11B-v3.0-Instruct} & \textbf{11} & \textbf{82.98} \\
  phi-4 & 15 & 81.71 \\
  Qwen2.5-7B & 7 & 74.60 \\
  Mistral-Nemo-Instruct-2407 & 12 & 74.14 \\
  cjvt/GaMS-9B-Instruct & 9 & 72.40 \\
  Apertus-8B-Instruct-2509 & 8 & 69.58 \\
  EuroLLM-9B-Instruct & 9 & 69.05 \\
  \underline{Bielik-11B-v2.6-Instruct} & \underline{11} & \underline{68.67} \\
  Apertus-8B-2509 & 8 & 59.04 \\
  \bottomrule
  \end{tabular}
  \caption{Belebele benchmark results showing model performance on reading comprehension across European languages (28 language subset).}
  \label{tab:belebele}
\end{table*}

On Belebele (Table~\ref{tab:belebele}), Bielik-11B-v3.0-Instruct achieves 82.98 average across European languages, representing a substantial improvement over the previous version Bielik-11B-v2.6-Instruct (68.67). This score places Bielik second overall, closely following Qwen2.5-14B-Instruct (85.91) and ahead of the larger phi-4 15B model (81.71). The strong performance on this benchmark confirms Bielik's robust reading comprehension abilities across multiple European languages.

\subsection{FLORES Machine Translation}

FLORES (FLORES-200) is a widely-used machine translation benchmark covering 200 languages, designed to evaluate translation quality across diverse linguistic families. The benchmark measures translation performance using BLEU scores, which assess n-gram overlap between model-generated translations and human reference translations. For Bielik evaluation, we assess translation performance across 20 European language pairs, focusing on bidirectional translations between Polish and other European languages, as well as translations among European languages. This evaluation provides insights into the model's multilingual translation capabilities across its target linguistic region. Results are shown in Table~\ref{tab:flores}.

\begin{table*}[t]
  \centering
  \small
  \begin{tabular}{lrrrr}
  \toprule
  \textbf{Model} & \textbf{Params (B)} & \textbf{AVG} & \textbf{to Polish} & \textbf{from Polish} \\
  \midrule
  EuroLLM-9B-Instruct$^{*}$ & 9 & 20.61 & 19.28 & 21.95 \\
  \textbf{Bielik-11B-v3.0-Instruct} & \textbf{11} & \textbf{19.22} & \textbf{18.54} & \textbf{19.91} \\
  \textbf{Bielik-11B-v3} & \textbf{11} & \textbf{17.85} & \textbf{17.60} & \textbf{18.11} \\
  phi-4 (15B) & 15 & 15.58 & 14.55 & 16.61 \\
  Mistral-Nemo-Instruct-2407 & 12 & 14.35 & 13.37 & 15.33 \\
  \underline{Bielik-11B-v2.6-Instruct} & \underline{11} & \underline{13.58} & \underline{15.77} & \underline{11.38} \\
  Qwen2.5-14B-Instruct & 14 & 13.24 & 12.55 & 13.93 \\
  \underline{Bielik-11B-v2} & \underline{11} & \underline{11.25} & \underline{14.86} & \underline{7.64} \\
  Qwen2.5-7B-Instruct & 7 & 11.34 & 10.43 & 12.26 \\
  \bottomrule
  \multicolumn{5}{l}{$^{*}$ EuroLLM was trained on FLORES dataset} \\
  \end{tabular}
  \caption{FLORES machine translation benchmark results showing translation performance across European languages (20 language pairs) measured by BLEU scores.}
  \label{tab:flores}
\end{table*}

On FLORES translation tasks (Table~\ref{tab:flores}), Bielik-11B-v3.0-Instruct achieves an average BLEU score of 19.22 across European language pairs, ranking second only to EuroLLM-9B-Instruct (20.61), which was trained on FLORES data. Notably, Bielik demonstrates balanced bidirectional translation capabilities with 18.54 BLEU for translation to Polish and 19.91 for translation from Polish. This represents a significant improvement over Bielik-11B-v2.6-Instruct (13.58 average), particularly in the from-Polish direction where v3.0 achieves 19.91 versus v2.6's 11.38. The base model Bielik-11B-v3 also shows strong translation performance (17.85 average), substantially outperforming larger models like phi-4 15B (15.58) and Qwen2.5-14B-Instruct (13.24).

\subsection{Summary of Evaluation Results}

The comprehensive evaluation demonstrates that Bielik-11B-v3.0-Instruct achieves state-of-the-art performance among open-source models of comparable size, with particularly exceptional results on Polish-specific benchmarks. The model excels in cultural knowledge (PLCC: 71.83\%, ranking first among all open-source models), general Polish language understanding (Open PL LLM Leaderboard: 65.93), and reading comprehension (Belebele: 82.98). Strong performance is also observed in specialized domains including medical knowledge (50.21\%), complex text understanding (CPTUB: 3.73), and emotional intelligence (EQ-Bench: 71.20).

Importantly, Bielik maintains competitive English language capabilities (Open LLM Leaderboard: 72.45) and demonstrates robust machine translation performance (FLORES: 19.22 BLEU), confirming its effectiveness as a multilingual European model. Across benchmarks, Bielik-11B-v3.0-Instruct frequently outperforms models with 2-6× more parameters, demonstrating exceptional parameter efficiency.

\section{Limitations and Biases}

The Bielik v3 series of models may produce factually incorrect output and should not be relied upon to generate completely accurate information in all contexts. These models were trained on diverse public datasets, and despite our extensive efforts to clean and filter the training data, they may occasionally generate content that is biased, offensive, or factually inaccurate. Users should apply appropriate caution and verification when deploying these models, particularly in sensitive or high-stakes applications.

\section{Conclusion}

We have presented Bielik 11B v3, a state-of-the-art multilingual language model optimized for Polish while maintaining strong capabilities across European languages. This work represents a significant advancement in developing high-performance language models for underrepresented languages through efficient scaling and comprehensive training methodologies.

Our main contributions include: (1) a depth up-scaling approach that achieves 11 billion parameters while remaining deployable on consumer-grade hardware; (2) a substantially expanded multilingual pre-training corpus spanning 32 languages with over 1.1 trillion tokens, representing a fivefold increase over our previous version; (3) a sophisticated four-stage training pipeline incorporating continuous pre-training, supervised fine-tuning, Direct Preference Optimization with the DPO-Positive variant, and reinforcement learning through Group Relative Policy Optimization; and (4) comprehensive evaluation across nine diverse benchmarks demonstrating exceptional performance on Polish-specific tasks while maintaining competitive multilingual capabilities.

The evaluation results demonstrate that Bielik 11B v3 achieves exceptional parameter efficiency, frequently outperforming models with 2-6 times more parameters. On Polish-specific benchmarks, the model establishes new standards: it ranks first among open-source models on the Polish Linguistic and Cultural Competency Benchmark (71.83\%), achieves 65.93 on the Open PL LLM Leaderboard, and scores 82.98 on Belebele reading comprehension across European languages. Notably, the model demonstrates balanced multilingual capabilities, scoring 64.8 on INCLUDE-base-44 across European languages while excelling on English benchmarks (72.45 average on Open LLM Leaderboard, with particularly strong mathematical reasoning performance of 85.60 on GSM8K).

Compared to the previous Bielik 11B v2 series, version 3 shows substantial improvements across most benchmarks, particularly in multilingual translation capabilities (FLORES: 19.22 vs. 11.25 BLEU) and reading comprehension (Belebele: 82.98 vs. 68.67). The model maintains competitive performance on specialized tasks including medical knowledge assessment (50.21\%) and complex text understanding (CPTUB: 3.73), demonstrating its versatility across diverse domains.

These results confirm that high-quality, resource-efficient language models can be developed for less-represented languages without sacrificing performance. The model's design prioritizes accessibility, with quantization support enabling deployment across diverse hardware configurations, from enterprise servers to consumer GPUs. This democratization of access is crucial for advancing AI capabilities beyond dominant languages.

Future research directions include exploring larger model variants to further push performance boundaries, extending language coverage to additional European and global languages, developing domain-specific adaptations for legal, medical, and scientific applications, and investigating multimodal extensions incorporating vision and speech capabilities. Additionally, we aim to conduct more extensive ablation studies to better understand the individual contributions of each training stage and to optimize the training pipeline further.

\section*{Acknowledgements}

We gratefully acknowledge Polish high-performance computing infrastructure PLGrid (HPC Center: ACK Cyfronet AGH) for providing computer facilities and support within computational grant no. PLG/2024/016951. 

The model could not have been created without the commitment and work of the entire SpeakLeash team, whose contribution is invaluable. Thanks to the hard work of many individuals, it was possible to gather a large amount of content in Polish and establish collaboration between the open-science SpeakLeash project and the HPC center: ACK Cyfronet AGH. Individuals who contributed to the creation of the model through their commitment to the open-science SpeakLeash project: Sebastian Kondracki, Marek Magry\'{s}, Igor Ciuciura, Szymon Baczy\'{n}ski, Dominika Basaj, Kuba So\l{}tys, Karol Jezierski, Jan Sowa, Anna Przyby\l{}, Agnieszka Ratajska, Witold Wydma\'{n}ski, Katarzyna Staros\l{}awska, Izabela Babis, Nina Babis, and many other wonderful researchers and enthusiasts of the AI world.

\bibliographystyle{cs-agh}
\bibliography{bibliography}

\end{document}